\documentclass[
]{ceurart}

\usepackage{listings}
\usepackage{changepage}
\usepackage{tikz}
\usetikzlibrary{positioning}
\usepackage{hyperref} 
\usepackage{tablefootnote}
\usepackage{color}
\begin{document}

\copyrightyear{2022}
\copyrightclause{Copyright for this paper by its authors.
  Use permitted under Creative Commons License Attribution 4.0
  International (CC BY 4.0).}

\conference{13th international workshop on Bibliometric-enhanced Information Retrieval (BIR 2023)\\
45th European Conference on Information Retrieval (ECIR 2023) in Dublin, Ireland}


\title{Text revision in Scientific Writing Assistance: An Overview }

\tnotemark[1]
\tnotetext[1]{You can use this document as the template for preparing your
  publication. We recommend using the latest version of the ceurart style.}

\author[1]{Léane Jourdan}[%
%orcid=,
%degree=,
%role=,
email=leane.jourdan@univ-nantes.fr
]
%\cormark[1]
\address[1]{Nantes Université, École Centrale Nantes, CNRS, LS2N, UMR 6004, F-44000 Nantes, France}
\author[1]{Florian Boudin}[%
%orcid=,
email=Florian.Boudin@univ-nantes.fr
]
%\address[2]{Vrije Universiteit Amsterdam, De Boelelaan 1105, 1081 HV Amsterdam, The Netherlands}
\author[1]{Richard Dufour}[%
%orcid=,
email=Richard.Dufour@univ-nantes.fr
%url=,
]
\author[1]{Nicolas Hernandez}[
email=Nicolas.Hernandez@univ-nantes.fr
]
%\cortext[1]{Corresponding author.}
%\fntext[1]{These authors contributed equally.}


\begin{abstract}
Writing a scientific article is a challenging task as it is a highly codified genre. Good writing skills are essential to properly convey ideas and results of research work. Since the majority of scientific articles are currently written in English, this exercise is all the more difficult for non-native English speakers as they additionally have to face language issues. 
This article aims to provide an overview of text revision in writing assistance in the scientific domain.
 We will examine the specificities of scientific writing, including the format and conventions commonly used in research articles. 
 Additionally, this overview will explore the various types of writing assistance tools available for text revision. Despite the evolution of the technology behind these tools through the years, from rule-based approaches to deep neural-based ones, challenges still exist (tools' accessibility, limited consideration of the context, inexplicit use of discursive information, etc.)

\end{abstract}


\maketitle
%\tableofcontents
\section{Introduction}
%Move 1: establishing a territory

    %%%Step 2: Making topic generalization
    The process of writing a scientific article can be complex and challenging, especially for junior researchers who often have to learn the conventions of scientific writing.
    This is even more true for researchers who are not native English speakers (ESL (English as Second Language) and EFL (English as foreign Language) learners) as strong writing skills are essential for effectively conveying ideas to the reader. 
    More generally, whether researchers are junior or senior, they must pay attention to the quality of writing in order to ensure that their work is correctly shared and understood by their audience.

    %%%Step 1: Claiming centrality
    To answer these needs, scientific writing assistance (SWA) has received more attention in recent years.
    %%%Step 3: Reviewing items of previous research
    In particular, a growing number of tools, language resources and events have emerged, aiming at helping scholars address these writing challenges.
    %def writing assistance
    SWA encompasses tools that answer to a range of different tasks, such as bibliographic management, text revision, spelling error correction or citation recommendation.
    %def text revision
    Considering that the field of SWA is vast, this paper focuses on summarizing approaches and tools for scientific text revision, defined as improving a draft on its content and phrasing to obtain the correct intended text in scientific style~\cite{du-etal-2022-read, alves2015progress}.
%Move 2: Establishing a niche
    %%% Step 1-B: Indicating a gap
%Move 3: Occupying the niche
    %%%Step 1A outlining purposes
    
    More specifically, we will cover only Natural Language Processing (NLP) tools, not language resources. However, the datasets used to train these tools will be mentioned. By conducting this overview, we hope to gain a better understanding of how these tools support scientific writers in effectively communicating their ideas and arguments and which approach they use for it.
    %%%Step 2: Announcing principal findings    
    %From this analysis we can see that even with all the tools currently available there are still challenges to face in terms of efficiency, functionalities or accessibility.
    Even with all the tools currently available, text revision in SWA is still an open field and our article tries to identify the challenges and future directions of research.
    %%%Step 3: Indicating RA structure
    
 The rest of this paper is structured as follows: In Section~\ref{sec:ecriture_scient}, we will first provide a definition of the research article genre and its characteristics. We will then define in Section~\ref{sec:aide_ecriture_scient} the task of text revision in scientific writing assistance. Following this, we will present an overview of the current tools that utilize NLP for this purpose. Finally, in Section~\ref{sec:challenges}, we will address the challenges that may be encountered in future research on this topic.

\section{Scientific writing}
\label{sec:ecriture_scient}

Scientific writing, also known as \textit{research writing}, is a subgenre of academic writing.
Among the literature, the definition of academic writing is ambiguous. It can be defined as a genre that encompasses all pieces of writing produced by students and researchers for academic work purposes in a university setting (essay, thesis, syllabus, etc.)~\cite{strobl2019digital}. However, scientific writing differs from academic writing in several aspects and has its own specificities and challenges.

Scientific writing is produced by researchers for other researchers, often in the form of research articles published in journals or conferences. It is expected to be concise, precise, clear, and follows a highly codified structure, tense / pronoun usage, and terminology~\cite{kallestinova2011write,bourekkache2022english}. The format style can be required by the targeted journal (for example IEEE style\footnote{\url{https://www.ieee.org/content/dam/ieee-org/ieee/web/org/conferences/style_references_manual.pdf}}).

Researchers, especially junior ones, often face particular difficulties when writing their first research articles, as they may lack experience with the codes, methodology, and techniques required in the scientific genre~\cite{bourekkache2022english}. This is a concern for  researchers across all domains, and it is especially relevant for those working outside their own discipline or in a multidisciplinary environment. Additionally, the majority of research articles today are written in English: for this reason, it is often ESL and EFL researchers who face the greatest difficulties in this regard as they need to learn the specificities of a foreign language at the same time. 

In this section, we will first present the structure followed by the majority of scientific articles, then describe the writing process of an article, and finally, show how the argumentative structure can be formalized.

\subsection{The structure of scientific articles}

The structure of a research article varies depending on the discipline and type of article (for example in the NLP domain: literature review, presentation of a corpus, creation of a new model, etc.). However, the commonly accepted structure for research articles is the IMRaD (or IMRD) model: \textit{Introduction, Methods, Results, and Discussion}. This structure was gradually adopted by the scientific community and became the most widely used pattern since the 1970s~\cite{swales_genre_analysis,sollaci2004introduction}. This model is the most popular and easiest to generalize across domains. To this model can be added common sections like \textit{Literature review/Related work} and \textit{Conclusion}.

The IMRaD format typically follows an hourglass pattern, beginning with a broad overview in the introduction and narrowing down to a specific focus on the motivation and goals of the research. The focus remains centered on this particular viewpoint throughout the related work, methods, and results sections. Finally, the paper broadens its scope again in the discussion and conclusion, considering the potential future directions and wider implications of the specific findings~\cite{swales_genre_analysis}.

Overall, the structure of a research article is designed to clearly and concisely present the research work. This structure helps to organize the information and makes it easy for readers to understand and evaluate the research. 
Each section has its own purposes that we describe here:
\begin{itemize}
    \item \textbf{Introduction}: The purposes of this section are to give context for the research, provide background information on the topic being studied and state the research question or hypothesis. The goal is to know: What question was studied? and why? 
    \item \textbf{Literature review/Related work}: This section discusses previous research on the topic or related domains. The goals are to know: What is the current state of the art? What are the gaps in the existing literature?
    \item \textbf{Methods}: This section describes the research design, including the participants, materials, and procedures used in the study. Its purpose is to answer the question: How was the problem studied?
    \item \textbf{Results}: This section presents the results of experiments, a score of a model on a task, etc. typically in the form of tables and figures. It answers the question: What were the findings of the study? 
    \item \textbf{Discussion}: This section interprets the results, discusses their implications, and suggests directions for future research. The purpose is to answer the questions: What do these findings mean? What do they imply?
    \item \textbf{Conclusion}: This section summarizes the key findings of the study and their significance. The conclusion must answer the question: What are the key elements the reader needs to remember from the paper?
\end{itemize}

\subsection{The writing process of scientific articles}
 Writing a research article is a process that has been studied in the learning analytics community usually broadening it to academic writing (including essays).
 
Different processes have been proposed throughout the years.
\cite{silveira2022guide} established one process to write a scientific article and~\cite{laksmi2006scaffolding,bailey2014academic} proposed processes for academic writing. All these propositions share similar steps that will be summarized in our considered process as \textit{Prewriting}, \textit{Drafting}, \textit{Revision} and \textit{Proof-reading}.

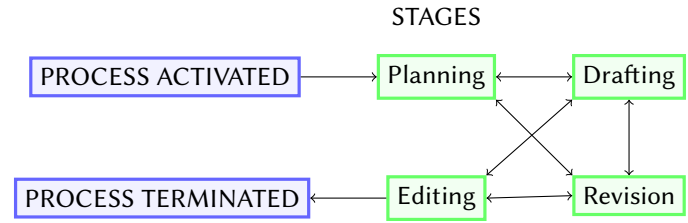
\begin{figure}
\centering
\begin{tikzpicture}[
roundnode/.style={circle, draw=green!60, fill=green!5, very thick, minimum size=7mm},
titlenode/.style={rectangle, draw=white!60, fill=white!5, very thick, minimum size=5mm},
bluerectangle/.style={rectangle, draw=blue!60, fill=blue!5, very thick, minimum size=5mm},
greenrectangle/.style={rectangle, draw=green!60, fill=green!5, very thick, minimum size=5mm}
]
%Nodes
\node[greenrectangle]      (step1)                              {Planning};
\node[greenrectangle]        (step4)       [below=of step1] {Editing};
\node[greenrectangle]      (step2)       [right=of step1] {Drafting};
\node[greenrectangle]        (step3)       [below=of step2] {Revision};
\node[bluerectangle]        (start)       [left=of step1] {PROCESS ACTIVATED};
\node[bluerectangle]        (finish)       [left=of step4] {PROCESS TERMINATED};
\node[titlenode, yshift=-22pt]        (title)       [above=of step1] {STAGES};
%Lines
\draw[->] (start.east) -- (step1.west);
\draw[<->] (step1.east) -- (step2.west);
\draw[<->] (step2.south) -- (step3.north);
\draw[<->] (step4.east) -- (step3.west);
\draw[<->] (step2.south west) -- (step4.north east);
\draw[<->] (step1.south east) -- (step3.north west);
\draw[<-] (finish.east) -- (step4.west);
\end{tikzpicture}
\caption{An example of a writing process proposed in~\cite{seow2002writing}} \label{fig:process_2}
\end{figure}

\cite{seow2002writing} proposed to add an iterative aspect and repetition of steps in this process as illustrated in Figure~\ref{fig:process_2}. This iterative notion can also be found in~\cite{du-etal-2022-read}, where they focus on the iterative aspect of the revision step.

Here is the writing process we will be considering, summarizing previous references and following the iterative pattern between steps from Figure~\ref{fig:process_2}:
\begin{itemize}
    \item \textbf{Step 1: Prewriting}
    \begin{itemize}
        \item Collect and organize ideas
        \item Write the outline
    \end{itemize}
    \item \textbf{Step 2: Drafting}
    \begin{itemize}
        \item Writing full sentences from notes
        \item Focus on content rather than form and structure
        \item Starting with the body in no particular order and next the introduction and conclusion
    \end{itemize}
    \item \textbf{Step 3: Revision}
    \begin{itemize}
        \item Changes in the structure of paragraphs and content of sentences
        \item Focusing on conciseness, clarity, connecting elements, and simplifying the text
        \item Making substantive rather than minor changes.
        \item Iteratively revised until the structure and phrasing is satisfying
        \item Correct grammar errors
    \end{itemize}
    \item \textbf{Step 4: Edition}
    \begin{itemize}
        \item Proofreading: Spelling error correction, minor changes, etc.
        \item Editing figures and tables
        \item Iteratively edit until no error is left
    \end{itemize}
\end{itemize}

This process can also be supported by research in the psycho-linguistic domain lead on expert writing summarized by~\cite{alves2015progress} (p.374) as: ``\emph{four cognitive processes support expert writing: \texttt{planning processes} that set rhetorical goals, which guide the generation and organization of ideas; \texttt{translating processes} that convert ideas into linguistic forms; \texttt{transcription processes} that draw on spelling and handwriting (or typing) to externalize language in the form of written text; and \texttt{revising processes} that monitor, evaluate, and change the intended and the actual written text}''. These definitions of the planning and revision tally with the previously described processes. The translating and transcription processes can be linked to the drafting step.

In our considered writing process, we are interested in the revision step. In this step, text coherence is particularly important. For this reason, it is essential to formalize the argumentative structure of scientific articles.

\subsection{Modelization of the argumentative structure}

In the discourse analysis and English for Academic Purposes research areas, efforts have been made to model the argumentative structure of scientific articles formally.
The argumentative structure shows the roles that argumentative discourse units (usually sentences) play in the overall argumentation~\cite{putra_teufel_tokunaga_2022,putra2021tiara}.

\begin{figure}[h]
    \centering
    \includegraphics[width=0.7\textwidth]{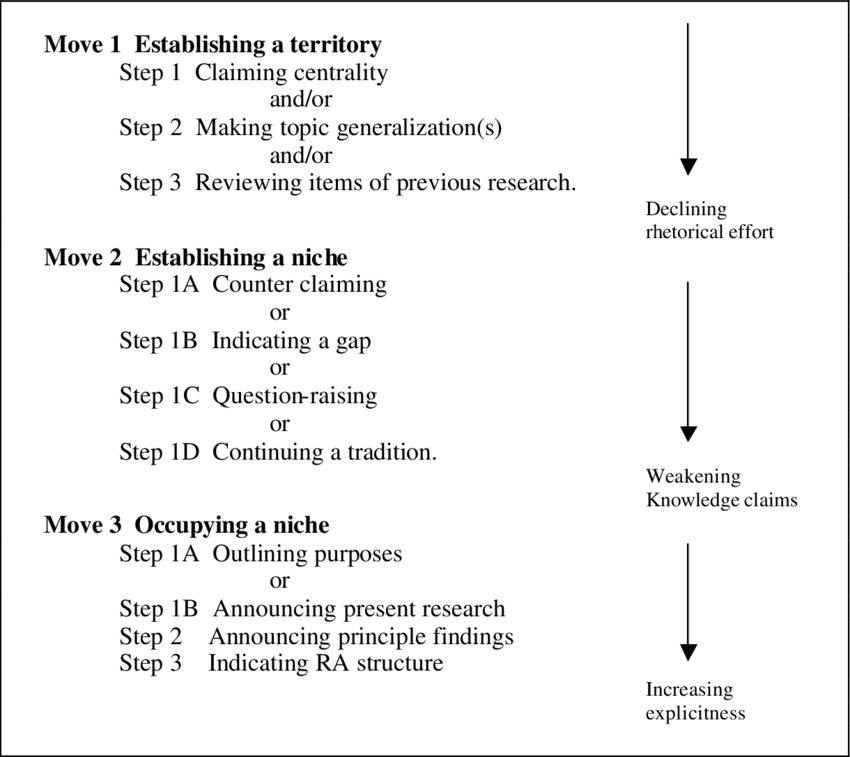}
    \caption{A CARS model for article introduction~\cite{swales_genre_analysis}.}
    \label{fig:cars_model}
\end{figure}

\cite{swales_genre_analysis} worked on the genre analysis of scientific articles and proposed 
the Creating a Research Space (CARS) model to describe the argumentative structure of the introduction of a research article. It is composed of four moves and eleven steps as illustrated in Figure \ref{fig:cars_model}.

An argumentative move is defined as a ``recurring and regularized communicative event''. It is a segment of text such as a phrase, sentence, or paragraph serving a specific purpose in discourse such as ``Indication of a gap'' in previous research~\cite{teufel1999annotation}(p.111)\cite{swales_genre_analysis}.
Each move is a series of functional strategies referred to as step~\cite{areyoureth}.

Derived from the CARS model and motivated by both a desire to propose a solution to analyze all sections of articles and the lack of resources on argumentative structure, several annotation schemes have been proposed.

%the first, the most famous
One of them is the Argumentative Zoning (AZ) model (or its improved version AZ-II~\cite{teufel2009towards}), based on~\cite{swales_genre_analysis} CARS model, that provides an analysis of the argumentative and rhetorical structure of a scientific paper~\cite{teufel2009towards}.
It is a sentence-level scheme used in the classification of sentences by their argumentative role within a scientific paper~\cite{teufel1999argumentative,liu2022incorporating,lawrence2020argument}. 
The categories, referred to as argumentative zones, are specific to the text type, in this case, research articles.

Other efforts have been made to propose different annotation schemes with their own focus. Where AZ focuses on how references are cited and for which purpose, schemes like Core-SC (Core Scientific Concepts) specifically designed for chemistry research articles focus on being easily understandable by humans~\cite{liakata2010corpora}.

\section{Text revision in scientific writing assistance}
\label{sec:aide_ecriture_scient}
Scientific writing assistance (SWA) encompasses a large range of tasks such as reference management, citation recommendation, grammatical error correction or sentence revision.
In this section, we are interested in the task of text revision and describing NLP tools that can be helpful for that task. 

\subsection{Definition of the text revision task}

Text revision is the task occurring at the revision step of the writing process. There is no single definition for what text revision in SWA is, but~\cite{du-etal-2022-read} (p.1) defined it as: ``\emph{identifying discrepancies between intended and instantiated text, deciding what edits to make, and how to make those desired edits}''.
In another work~\cite{li2022text} (p.58), it is defined as ``\emph{a series of text generation tasks including but not limited to style transfer, text simplification, counterfactual debiasing, grammar error correction and argument reframing}''.
Text revision is the transformation of an input text into an improved version fitting a desired attribute (formality, clarity, etc.), closer to the intended text.
In recent work, the text revision task is often limited to sentences.
The sentence revision task, also called SentRev, is defined by~\cite{ito-etal-2019-diamonds} (p.41) as ``\emph{revising and editing incomplete draft sentences to create final versions}''.

No events have been organized specifically on text revision but the field of SWA has seen a growing interest in recent years. For example, the Helping our own (HOO) shared task, held in 2011, aimed to develop and evaluate automated tools and techniques for error correction that can assist authors in their writing focusing on the NLP community~\cite{dale-kilgarriff-2011-helping}. Years later, the Intelligent and Interactive Writing Assistant (In2Writing\footnote{\url{https://in2writing.glitch.me/}}) workshop began in 2022 with the goal of ``\emph{facilitate discussion around writing assistants, thereby enhancing our understanding of their usage in the writing process and predicting the consequences}''. These examples highlight the research efforts and the resurging interest in the field of writing assistance.

 Below are the text revision tools that will be considered in our study. They are grouped into three general categories depending on whether they suggest modifications, correct only grammar and spelling errors, or only annotate the text to visualize its structure.
\begin{itemize}
    \item \textbf{Sentence revision tools:}
    These tools provide automatic suggestions for the revision step of the writing process (changing the structure of a sentence, rephrasing for clarity, etc.) at sentence level. Sentence revision is iterative~\cite{du-etal-2022-read} and 1-to-N~\cite{ito-etal-2019-diamonds} as a sentence can have several correct revisions.
    \item \textbf{Grammar checkers:} These tools tackle grammar error correction (GEC) and spelling error correction (SEC) tasks as they can detect grammar, spelling, and punctuation errors in a document and propose a correction automatically. Some can be built directly into word processing software. They are considered a part of the text revision step as making a grammar change can substantively modify the sentence.
    \item \textbf{Moves annotators:} These tools are dedicated to academic writing. They highlight the moves (from CARS model or another framework of moves), for example by color-coded them. It will make the argumentative structure visible in order to help the writer revise and correct their draft.
    These tools can also give suggestions on the order of the moves or if there is any missing one.
\end{itemize}

\subsection{Currently available tools}
\label{sec:outils}

There are a number of writing assistance tools acting on text revision.
In this section, we will present a selection of some pertinent tools that exist and can be applied for scientific writing. Table~\ref{table:outils} summarizes their main characteristics. 

With the exception of ChatGPT, the tools presented in this section are currently limited to the English language.

\begin{table}[h]
\begin{adjustwidth}{-2.2cm}{}
\begin{tabular}{|c|c|c|c|c|c|c|c| }
\hline
    \textbf{Category} &\textbf{Tool} & \textbf{Year\tablefootnote{The year corresponds to the last known update.}} & \textbf{Domain}  & \textbf{Approach} & \textbf{Task}  & \textbf{Availability}\\
\hline
&Langsmith~\cite{ito2020langsmith} & 2020 & Scientific & Transformer based & SentRev, text completion, & Free and paid plans \\
Sentence && & & & GEC and SEC & \\
\cline{2-7}
revision &R3~\cite{du-etal-2022-read} & 2022 & General/  & Transformer based & SentRev & Open source \\
 tools& & & Scientific & & & \\
\cline{2-7}
&Chat GPT~\cite{ouyang2022training} & 2023 & General & GPT 3.5 & Text generation & Free and paid plans \\
\hline
&Grammarly\tablefootnote{\url{https://www.grammarly.com}} & 2023 & General & Transformer based  & Correctness, clarity,  &  Free and paid plans \\
Grammar&& & & &  engagement, delivery & \\
\cline{2-7}
checkers&LinggleWrite~\cite{tsai-etal-2020-lingglewrite} & 2020 & Academic & LSTM/ Bi-LSTM & Suggestions, & Free to use \\ 
&& & & & essay scoring & \\
\hline 
&Mover~\cite{anthony2003mover} & 2016 & Academic &Naive Bayes & Moves analysis & Free to use \\
Move& &  &  &Classifier&  & \\
\cline{2-7}
annotators&RWT~\cite{cotos2016computer} & - & Academic & Probabilistic models & Moves analysis & Limited access\\
\cline{2-7}
&AcaWriter~\cite{knight2020acawriter} & 2022 & Academic & Rule-based & Moves analysis & Open source \\
\hline

 \end{tabular}
 \caption{Description of the tools currently available for text revision.}
  
\end{adjustwidth}
\label{table:outils}
\end{table}

\subsubsection{Sentence revision tools}

In this section, we will present \textit{Langsmith} and \textit{R3}, two sentence revision tools trained on scientific articles. We will also discuss \textit{ChatGPT}, one of the most advanced question-answer NLP tools, that can be used for text revision.

\paragraph*{Langsmith}

is an interactive academic sentence revision system developed by a team of researchers from Tohoku University, Edge Intelligence Systems Inc. and RIKEN~\cite{ito2020langsmith}. The system was released in 2020 and is available online through the \href{https://editor.langsmith.co.jp/}{Langsmith editor}\footnote{\url{https://editor.langsmith.co.jp/}}  with free or paid plan options.
\textit{Langsmith} is designed to be used on NLP research papers enabling domain-specific revisions such as correcting technical terms and is mainly targeted at inexperienced and non-native researchers. The system's main feature is its revision function, but it also includes a text completion and an error correction feature, two tasks that can be considered part of the revision step in our writing process.
The revision function can suggest fluent, academic-style sentences to writers based on their rough, incomplete phrases or sentences.

Their system for revision uses an encoder-decoder with a convolution module and is trained on synthetic data they created by altering sentences from research articles.
The text completion feature is specialized in academic writing and leverage a GPT-2 small model fined-tuned on papers from the ACL Anthology.

With \textit{Langsmith}, users can request a specific revision or select one from several candidates. This work pointed out the importance of considering the 1-to-N nature of the revision task, as one sentence can have several correct revised versions.

\paragraph{R3} (Read, Revise, and Repeat) is a human-in-the-loop model for iterative text revision proposed in 2022~\cite{du-etal-2022-read}. The code is available on Github\footnote{\url{https://github.com/vipulraheja/IteraTeR/}}.
It is composed of a fine-tuned RoBERTa-large and a fine-tuned PEGASUS-LARGE~\cite{du-etal-2022-read}. 
The training data for \textit{R3} (dataset \texttt{IteraTeR}) was collected from text revision data across three domains: ArXiv, Wikipedia, and Wikinews.
The model is designed to be a general revision model, but as it has been trained on research articles, it can be considered a scientific writing assistant.

 \textit{R3} is an interface where the writer can upload a document and then iteratively accept or refuse sets of revisions, as illustrated in this \href{https://youtu.be/lK08tIpEoaE}{video}\footnote{\url{https://youtu.be/lK08tIpEoaE}}. 
 The main advantage of \textit{R3} is that it offers a new way of thinking about the task of text revision by introducing the importance of both the iterative aspect of the process and adding human in the loop. However, even presented as {\it text revision}, it actually processes sentences one by one independently making it a model for the \textit{SentRev} task.

\paragraph{ChatGPT}
is a large-scale generative language model developed by OpenAI and launched on November 30, 2022.
It is based on GPT-3.5 and has been fine-tuned for dialogue, allowing it to interact in a conversational manner. 
 The first version of the chatbot was trained on a dataset of human-human conversations, where one human plays the role of the chatbot. Then, the model has been fine-tuned through reinforcement learning with human AI trainers ranking samples of the chatbot's responses in simulated conversations.
 
 \textit{ChatGPT} can be accessed through a web application, where users can type in their questions or requests for assistance in a chat interface. The model is not specifically tailored for scientific or academic writing. However, it can be applied to a variety of tasks related to scientific writing, such as text revision, simplification and restructuration, translation, grammatical error correction (GEC), etc.
 Additionally, it is available in a range of different languages.

It is a new actor in writing assistance and can be used for scientific writing by asking specific queries. Here are some ideas of prompts to use it as an assistant in your writing such as: 
\begin{itemize}
    \item \textit{Can you revise and correct this in an academic style? : ``\texttt{<your draft>}''} 
    \item \textit{Can you revise the abstract I wrote for my paper on \texttt{<subject of your paper>}: ``\texttt{<your abstract>}''}
    \item \textit{Can you translate this paragraph where I talk about \texttt{<subject and purpose of your paragraph>}: ``\texttt{<your paragraph>}''.}
    \item \textit{Can you rephrase this paragraph making it more \texttt{<add your criteria>}: ``\texttt{<your abstract>}''}
\end{itemize}
All these prompts follow the same pattern with a different intent: \textit{\texttt{<Your instructions for the model>}: ``\texttt{<Your draft version>}''}. Giving more context in your prompt will often lead to better results.

\textit{ChatGPT} is currently in beta and free of use but this may change in the future with the creation of a paid plan.
However, it should be noted that it still has some limitations, including sensitivity to phrasing in the prompt and the potential for plausible-sounding but incorrect or nonsensical answers. 

Moreover, since its launch, \textit{ChatGPT} has been highly criticized as its use raises questions regarding ethics and plagiarism. In January, ACL 2023 released a post on their blog\footnote{\url{https://2023.aclweb.org/blog/ACL-2023-policy/}} regarding their policy on AI Writing Assistance focusing on the use of \textit{ChatGPT}. In this post, they discourage its use to produce new ideas and text. They ask the authors to acknowledge the use of writing assistance tools except when using them purely with the language of the paper.

\subsubsection{Grammar checkers}
In this section, various error-checking tools will be discussed. These tools are commonly utilized for general writing and include \textit{Grammarly}, {\it LanguageTool}, {\it Ginger}, etc. Each tool offers a range of functionalities, including correcting spelling and grammar errors, detecting paraphrasing, essay scoring, etc. As an extensive number of these tools exist, only two will be presented here.
It should be noted that these tools do not take into account the argumentative structure of the text nor a large context.

\paragraph{Grammarly}
is one of the most widely and easy-to-use error-checking tool. It was developed by Grammarly Inc. in 2009 and is currently available as an online text editor, as well as a mobile app and browser extension. The browser extension also allows for integration with other popular text editors, such as Overleaf and Google Docs.
 
 The free version offers suggestions in correctness (spelling, grammar, or punctuation) and clarity, and their online editor allows one to set specific goals for the writing in terms of targeted audience, formality, and intent. The premium offer extends these features with suggestions in engagement and delivery and an additional goal to specify the domain (where ``academic'' is one of the available options).

\paragraph{LinggleWrite}

is a writing coach for essay writing targeted to English learners providing writing suggestions on an input text. It is derived from \textit{Linggle- Language Reference Search Engines} and was released in 2020 by NLPLab~\cite{tsai-etal-2020-lingglewrite}.
Unlike some other tools, it is only available as a web application and there is no existing browser extension.

The tool is specifically designed for essay writing in an academic setting and was trained using the EF-Cambridge Open Language Database and the First Certificate in English dataset.
The system behind it consists of four components: writing suggestions, essay scoring, GEC, and corrective feedback~\cite{tsai-etal-2020-lingglewrite}. The writing suggestions are based on a dictionary of grammatical patterns, hand built or extracted from a corpus.
The model behind \textit{LinggleWrite} utilizes a combination of LSTM with an attention layer for essay scoring and BiLSTM-CRF, BERT, and Flair embeddings for GEC. 

\subsubsection{Move annotators}

%writing tools for university  
\textit{Mover}, {\it Research Writing Tutor}, and \textit{AcaWriter} are tools from the writing analytics domain, which is a sub-domain of learning analytics~\cite{areyoureth}.
These tools are rhetorical moves annotators and writing feedback tools. Their goal is to provide formative feedback to students by automatically identifying the argumentative structure of their text. Move annotators purpose is to improve students' writing capacities and help them revise their drafts through the visualization of the structure.
These tools are more extensively described in~\cite{areyoureth}.

Move annotators are usually academic tools and currently present two limitations.
First, moves annotators are often linked to Universities and their access is secured by a password.
Secondly, in the argumentative structure and moves literature, the introduction and the abstract have been the central point of focus as it contains discourse information about the subjects and content of the article. This part contains much more rhetorical information but is one of the last to write in the writing process. There is a lack of research and proposition from move framework formalizing the other parts of research articles.

\paragraph{Mover}

is a text structure (moves) analysis software developed by Laurence Anthony and George V. Lashkia in 2003. This software is intended to assist students in evaluating and revising their drafts~\cite{areyoureth}.
The algorithm employed by \textit{Mover} is a Naive Bayes classifier approach and was trained on a supervised learning task on a corpus of 100 research abstracts in the field of information technology~\cite{anthony2003mover,areyoureth}. The moves model utilized in the annotation process is the ``Modified (CARS) Model'' by~\cite{anthony1999writing}. 
The software can be downloaded on Laurence Anthony's website\footnote{\url{https://www.laurenceanthony.net/software/antmover/}}.

\paragraph{Research Writing Tutor (RWT)}

is a web-based application developed by Elena Cotos and Stephen Gilbert for academic writing assistance. 
It is composed of three modules, one of which provides feedback and analysis of written text. 
This module identifies the rhetorical structure and employs an extended move/step framework of the CARS model to color-code the structure for better visualization~\cite{cotos2016computer} across all sections of the IMRaD structure, comprising a total of 61 steps distributed in 14 moves.

Based on this structural analysis, \textit{RWT} will provide feedback~\cite{areyoureth} on the use of moves, comparing the draft's moves distribution to a goal distribution extracted from articles in the student's discipline domain~\cite{cotos2016computer}. 
Additionally, it will analyze the use of steps in the form of comments and clarifying questions about the rhetorical intent of a given sentence~\cite{cotos2016computer}.
The distributions are extracted from 900 introductions in 30 disciplines~\cite{cotos2020understanding}.
 
For this supervised classification problem task, \textit{RWT} uses probabilistic language models~\cite{cotos2015furthering}. 
Currently, access to \textit{RWT} is restricted to Iowa State University, however, guest accounts can be created upon request~\cite{cotos2016computer}.

\paragraph{AcaWriter}
is a  web-based application that is part of the Academic Writing Analytics (AWA) project by the University of Technology Sidney (UTS).
\textit{AcaWriter} uses the \textit{rhetorically salient sentences} as a moves framework to label the sentences using a rule-based system~\cite{areyoureth,knight2020acawriter}.
When used for abstracts and introductions, it provides feedback on the order of the moves or if there is any missing one.

\textit{AcaWriter} is accessible to UTS staff and students and a demo version is available for external users. They also propose an open-source platform for institutions that wish to host their own version of \textit{AcaWriter}\footnote{\url{https://cic.uts.edu.au/tools/awa/}}.

\section{Conclusion and future directions}
\label{sec:challenges}

In this work, we highlighted the unique characteristics of scientific article writing as a highly codified genre.
The type of assistance needed for writing an article varies depending on the task. We were particularly interested in the text revision task.
We presented an overview of currently available writing assistance tools for this task and how they address it.
However, all existing writing assistance tools have not been covered as they are too numerous, and only a few representative ones have been selected.

From our analysis of existing tools, we identify seven major challenges to face in future work:
\begin{enumerate}

    \item \textbf{Benchmarking performance:} While our article presented a selection of tools, proposed a classification of these writing assistants and identified their available features, comparing the tools' performances is still an open issue. Finally, quite a few evaluations accompany the tools developed and it is currently impossible to compare their performances.
    
    \item \textbf{Considering a larger context:} Currently, sentences are treated independently. Considering a larger context (for example revision at paragraph level) could be beneficial for the text revision task. One potential solution for this is to look into other domains, such as machine translation, to see how current models are trained to consider larger contexts~\cite{Majumder2022,li-etal-2020-multi-encoder,feng-etal-2022-learn,chen-etal-2020-modeling}.
    
    \item \textbf{Taking discursive analysis into consideration:} Discourse analysis allows annotating the organization of discourse linking information in the text~\cite{taboada2006rhetorical,danlos-2011-analyse}. Considering it would permit catching long-distance dependencies in the text which is essential for a good organization of scientific documents.

    \item \textbf{Including the argumentative structure:} Current tools usually only label each sentence to highlight the structure and give some feedback. However, there is a lack of guidance on how to effectively structure an argument and present evidence to support claims. One research direction would be to use argument mining techniques to study the relationship between arguments~\cite{liu2022incorporating,lawrence2020argument}. 

    \item \textbf{Lack of available resources:} Existing corpora for the revision task, such as \texttt{IteraTeR}~\cite{du-etal-2022-read} and \texttt{arXivEdits}~\cite{jiang2022arxivedits}, are composed of final articles and their versions before revisions, collected from arXiv. However, an issue with arXiv as a source of data is that the first versions of submitted articles have already been proofread, sometimes even revised by peers.
    Although there is no simple solution to that data problem, one possibility would be to ask researchers to contribute towards building such resources by providing early drafts of accepted papers.

    \item \textbf{Improving accessibility and transparency:} Some writing tools are not publicly accessible while others are not properly described in the research literature. The accessibility issue appears mostly with academic tools, some exist inside universities with access limited (non-commercial use) to students and staff~\cite{strobl2019digital}. The transparency issue occurs mainly with lucrative tools proposed by private companies as they do not always publish a paper on how they build and train their models nor share their training data.

    \item \textbf{Emergence of ethical issues:} The use of text revision tools raises some ethical issues about plagiarism, intellectual property and the potential impact on the quality and integrity of research. Additionally, it is worth examining the impact of these tools on English learners, as they may facilitate the writing process to such an extent that proper writing skills and language knowledge are not developed~\cite{SHINTANI2013286,SAMPSON2012494,chen2016efl}.
 
\end{enumerate}

\bibliography{biblio}

\end{document}